\documentclass[conference]{IEEEtran}
\IEEEoverridecommandlockouts
\usepackage{cite}
\usepackage{amsmath,amssymb,amsfonts}
\usepackage{algorithmic}
\usepackage{graphicx}
\usepackage{textcomp}
\usepackage{xcolor}
\usepackage[colorlinks=true, citecolor=blue]{hyperref}

\def\BibTeX{{\rm B\kern-.05em{\sc i\kern-.025em b}\kern-.08em
    T\kern-.1667em\lower.7ex\hbox{E}\kern-.125emX}}
    
\begin{document}

\title{Evaluating Rank-N-Contrast: Continuous and Robust Representations for Regression \\
}

\author{\IEEEauthorblockN{Alexandre Chidiac}
\IEEEauthorblockA{\textit{Georgia Institute of Technology}\\
\textit{Télécom SudParis}\\
achidiac3@gatech.edu\\}
\and
\IEEEauthorblockN{Valentin Six}
\IEEEauthorblockA{\textit{Georgia Institute of Technology}\\
\textit{Télécom SudParis}\\
vsix3@gatech.edu\\}
\and
\IEEEauthorblockN{Arkin Worlikar}
\IEEEauthorblockA{\textit{Georgia Institute of Technology}\\
aworlikar6@gatech.edu}
}

\maketitle

\begin{abstract}
This document is an evaluation of the original "Rank-N-Contrast" \cite{zha2024rank} paper published in 2023. This evaluation is done for academic purposes. It is often difficult for deep regression models to capture the continuous nature of sample orders, creating fragmented representations and suboptimal performance. To address this, we reproduced the Rank-N-Contrast (RNC) framework, which learns continuous representations by contrasting samples by their rankings in the target space. Our study validates RNC's theoretical and empirical benefits, including improved performance and robustness. We extended the evaluation to an additional regression dataset and conducted robustness tests using a holdout method, where a specific range of continuous data was excluded from the training set. This approach assessed the model's ability to generalize to unseen data and achieve state-of-the-art performance. This replication study validates the original findings and broadens the understanding of RNC's applicability and robustness.\\
\end{abstract}


\section{Introduction}

Regression problems are fundamental and can be encountered across various domains, including estimating age from human appearance \cite{rothe2015dex}. The continuity inherent in regression targets necessitates models that can capture and predict these continuous relationships effectively. Existing regression methods mostly focus on predicting target values directly, and use classical distance-based loss functions to evaluate the predictions of the regression model. However, these approaches often neglect the importance of the learned representation of the training data, which is crucial for capturing the underlying continuous relationships in regression tasks.
Contrastive learning is a technique used to better understand data by comparing different examples and assessing their similarity. This technique has demonstrated large success in representation learning for classification and segmentation tasks. However, applying contrastive learning to regression tasks has been relatively unexplored. Most representation learning methods, such as supervised SupCon \cite{khosla2020supervised}, often overlook the continuous nature of data and fail to capture the intrinsic continuity in the data.\\

To address these limitations, a new solution for generic regression learning is introduced : Rank-N-Contrast (RNC). This framework first learns a regression-aware representation that orders the distances in the embedding space based on the target values, and then leverages this representation to predict continuous targets. The Rank-N-Contrast loss ($L_{RNC}$) is introduced, which ranks the samples in a batch according to their labels and contrasts them based on their relative rankings. It has been proved that optimizing $L_{RNC}$ leads to improved performance for regression tasks, by ordering features according to the continuous labels. RNC learns continuous representations that capture the intrinsic ordered relationships between samples. The RNC framework can easily be used with existing regression methods in order to map the learned representation to the final prediction values.\\

The authors also benchmark RNC practically against state-of-the-art (SOTA) regression and representation learning schemes on five regression datasets. The experiments verify the better performance, robustness, and efficiency of RNC in learning continuous targets. \\

The overall contributions of the paper are the following : 

- We identify the limitations of current regression and representation learning methods for continuous targets.
- We explain RNC, a method that learns continuous representations for regression.
- We experiment on a real-world regression dataset and check the superior performance of RNC compared to SOTA.
- We confirm other properties of RNC for data efficiency, robustness and data corruptions.

\section{State of the art}

\subsection{Regression Learning}

Many different techniques have been employed to improve the performance of models on regression tasks. The usual approach focuses on the loss between the prediction and the target value, but other options have been explored. A way of turning the regression task into a classification task is to divide the continuous range of values into bins \cite{rothe2015dex}. A variant to this approach can be performed by combining the predictions of many ordinal classifiers \cite{niu2016ordinal}. Some work on the regularization of the embedding space. 
The presented work \cite{zha2024rank}, in contrast to existing methods and classical approaches, focuses on providing a regression-aware representation learning method, allowing better performance on regression task. It is important to note that RNC is compatible with already-existing regression methods.

\subsection{Representation Learning}

State-of-the-art representation learning techniques include SupCon (SUPervised CONtrastive Learning) \cite{khosla2020supervised}, DINO (self-DIstillation with NO labels)\cite{caron2021emerging}, and SimCLR (SIMple framework for Contrastive Learning of visual Representations) \cite{chen2020simple}.

SupCon is a method using contrastive learning for classification. Positive pairs are generated from the same class, and negative pairs consist of samples from differing classes. This technique can be adapted to a regression problem by splitting the regression space into bins, which are treated as classes \cite{khosla2020supervised}. In this way, state of the art Contrastive Learning techniques for classification, can be adapted for regression and compared to the Rank-N-Contrast method.
DINO is a method which is based on the self-distillation framework, which has a "student" and "teacher" network. The representation is learned by maximizing the similarity of features learned by both networks.
SimCLR is another contrastive learning method which generates positive pairs by augmenting a sample in two different ways. While negative pairs are generated from different samples. Our works differ from previous papers as it uses label continuity for designing a representation learning framework adapted to regression tasks.

\section{Problem Definition}
The problem we are addressing is improving the accuracy of a regression task using representation learning, with the specific new application of estimating weight from mango images. Traditional regression methods often struggle with complex, high-dimensional data like images, as they rely on manually selected features or simplistic representations of the data. By employing representation learning, we aim to automatically learn meaningful features directly from the raw images, which should lead to more accurate predictions. 

Our hypothesis, following the methodology proposed in the paper we are replicating, is that using the Rank-N-Contrast loss function $L_{RNC}$ \cite{zha2024rank} will improve the model’s ability to differentiate small differences in data that correspond to differences in age, and therefore, enhance overall accuracy. By replicating this approach, our goal is to evaluate its effectiveness and robustness on a different dataset, while potentially identifying any limitations or challenges in its application.

\subsection{Background}
Here are some important concepts that are essential to the understanding of the paper:

\begin{itemize}
    \item \textbf{Representation Learning and Encoder:} The goal of representation learning is to automatically discover the most useful features or representations from raw data. In practice, this is achieved by optimizing a function called an encoder, which extracts a relevant feature embedding from the original input.
    
    \item \textbf{Regression and Predictor:} The task at hand is to predict a continuous variable (like weight) from the mango images, which makes this a regression problem. Unlike classification, where the goal is to assign labels, here we need to predict a value that can vary continuously. We learn a function called a predictor to estimate the best value for a given input. Regression tasks often involve minimizing a loss function.
    
    \item \textbf{Contrastive Learning:} Contrastive learning typically involves pushing apart representations that are dissimilar while associating those that are similar. This approach is traditionally used for classification tasks, but we will use it for regression. The Rank-N-Contrast loss function, $L_{RNC}$, ensures that the model not only learns accurate representations, but also that these representations reflect the relative ordering of continuous labels like age or weight.
    
\end{itemize}

\subsection{Mathematical Concepts}

The loss function $L_{RNC}$ ensures that the model not only learns accurate representations, but also that these representations reflect the relative ordering of continuous labels like age (\cite{rothe2015dex, zha2024rank}).

The task is to estimate a function $f: X \to R^{d_e}$, where:

\begin{itemize}
    \item $X$ is the input space, in this case, images.
    \item $R^{d_e}$ is the $d_e$-dimensional latent space (i.e., the feature space), where each input is mapped to a vector representation by $f$.
    \item $g: R^{d_e} \to R^d$ is the final regression layer that predicts the target continuous variable based on the features extracted by $f$.
\end{itemize}

\subsubsection{\texorpdfstring{$L_{RNC}$ and $f$}{L RNC and f}}
By ranking and contrasting the embeddings, for each anchor sample, the RNC loss ensures that its feature embedding is more similar to those of samples with closer labels and less similar to those with more distant labels.

\subsubsection{Likelihood and Estimating \texorpdfstring{$f$}{f}}
The likelihood of any sample $v_j$, given an anchor $v_i$, is modeled based on the similarity of their feature embeddings:
\[
P(v_j \mid v_i, S_{i,j}) = \frac{\exp(\text{sim}(v_i, v_j) / \tau)}{\sum_{k \in S_{i,j}} \exp(\text{sim}(v_i, v_k) / \tau)}
\]
where $\text{sim}(v_i, v_j)$ measures the similarity between embeddings, and $S_{i,j}$ represents a set of samples ranked higher than $v_j$ based on their label distance.

\subsubsection{Estimation of \texorpdfstring{$f$}{f}}
$L_{RNC}$ helps optimize $f$ by pushing the model to organize the feature space according to the relative distances between continuous labels. The per-sample $L_{RNC}$ $\ell_{RNC}(i)$ is minimized to adjust $f$ in such a way that the resulting embeddings reflect the best possible rank ordering:
\[
L_{RNC} = \frac{1}{2N} \sum_{i=1}^{2N} \ell_{RNC}(i)
\]
By minimizing this loss, we improve the estimation of $f$.\\
Once we have a good representation, we can perform the prediction step using classical regression on the encoded data.

\section{Evaluation}

\subsection{Implementation}

In order to implement the theoretical and mathematical concepts to conduct our experiments, we will use and adapt the original code used in the original paper "Rank-N-Contrast: Learning Continuous Representations for Regression" \cite{zha2024rank}. This will allow us to have an easier and direct implementation of the previously explained concepts, like the $L_{RNC}$ loss function. We will also need to obtain the dataset with the mango fruit images \cite{ismail2022estimating}. Using the provided code and the downloaded data, we can train different models on our data to compare their performance ; more specifically, the models that will be put to the test have a different loss function (standard $L_1$ or $L_{RNC}$) for the representation learning step. This will enable us to compare the performance and impact of our RNC framework for representation learning compared to other SOTA representation learning methods.

\subsection{Evaluation of methodology}
As part of evaluating the robustness of the methodology, we will introduce various experimental variations designed to "break" the model:

\subsubsection{Removing a Certain Range of Ages}
By selectively removing data corresponding to certain age groups in the initial dataset, we can test how well the model generalizes across unseen age ranges and identify potential biases in the learned representations. This will help us understand whether the model has learned to extrapolate for missing age groups, or if it struggles with unseen data.

\subsubsection{Testing with Different Dataset}
We will test the performance of our models on a different and smaller dataset, to see the impact of dataset size.

\subsubsection{Use of Various Metrics}
As for most of the regression tasks, loss and accuracy metrics will be used to assess the model.

\section{Potential challenges}
The main challenge when it comes to replicating a paper is to achieve an in-depth understanding of the theory and concepts exposed. Without this, it's difficult to effectively reproduce the results or adapt the methodology to new datasets or applications.
For instance, in our case of replicating the Rank-N-Contrast (RNC) approach for improving regression tasks, the key challenge lies in fully comprehending the theoretical basis of the loss function and its relationship to contrastive learning. \\

\section{Results}

In this section we test models that differ mainly in the loss function used for representation learning at the encoder:  "$L_1$" refers to a model that was trained using $L_1$ for both the representation and the prediction steps ; "$L_{RNC}+L_1$" refers to a model that used the RNC loss function ($L_{RNC}$) at the encoder and standard $L_1$ loss for the prediction.\\

The first experiment was about reproducing the achieved results in the original paper, by comparing the performance on a regression task of a model trained using the standard $L_1$ loss compared using the $L_{RNC}$ for the embedding space. For that, we used a known dataset from the original paper, AgeDB \cite{moschoglou2017agedb}, used for age estimation from images of faces. First, we used this dataset to train a model using the $L_1$ loss only for both the embedding space and the final prediction ; then we trained the encoder for the embedding space alone using the RNC loss introduced earlier and also trained the whole model with $L_1$, using the trained encoder. We obtain two trained regression models that we can for which we can compare the performances and see the impact of using $L_{RNC}$ at the embedding step. \\

\begin{figure}[htbp]
\centering
\includegraphics[width=3.5in]{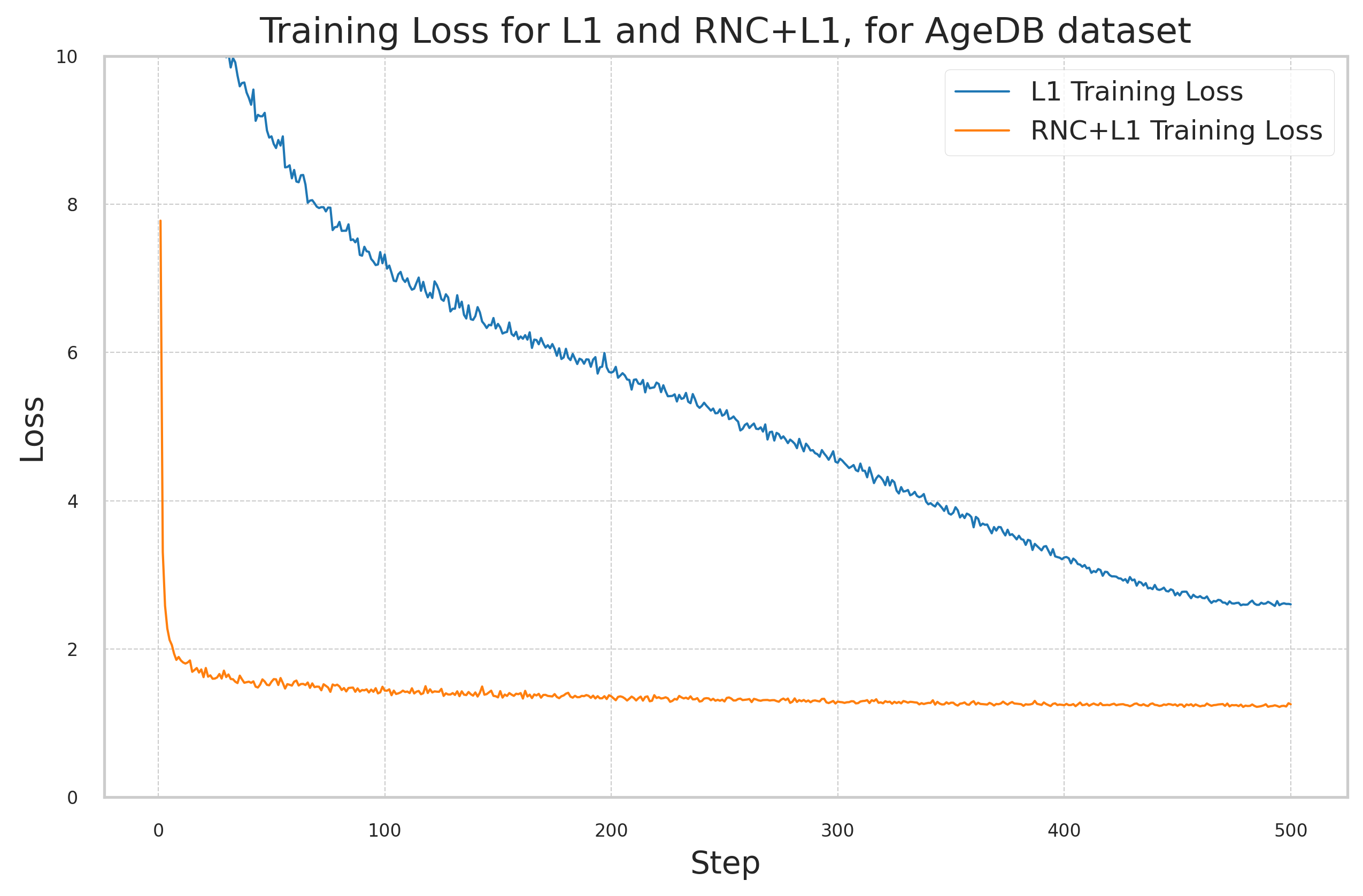}
\caption{Training Loss for $L_1$ and $L_{RNC}+L_1$ on AgeDB dataset}
\label{fig1}
\end{figure}

Looking at Figure \ref{fig1}, we can see that for $L_1$, the value of the loss decreases smoothly and seems to converge to a value of 2 after around 500 epochs. For the $L_{RNC}+L_1$, where we first train the encoder with $L_{RNC}$, the loss goes down very rapidly and converges to around 1, which is half of the final loss for the other model. Looking at those results, we can see that for the AgeDB dataset, the model for which the encoder was trained using $L_{RNC}$ achieves a smaller loss value and converges faster.\\

\begin{figure}[htbp]
\centering
\includegraphics[width=3.5in]{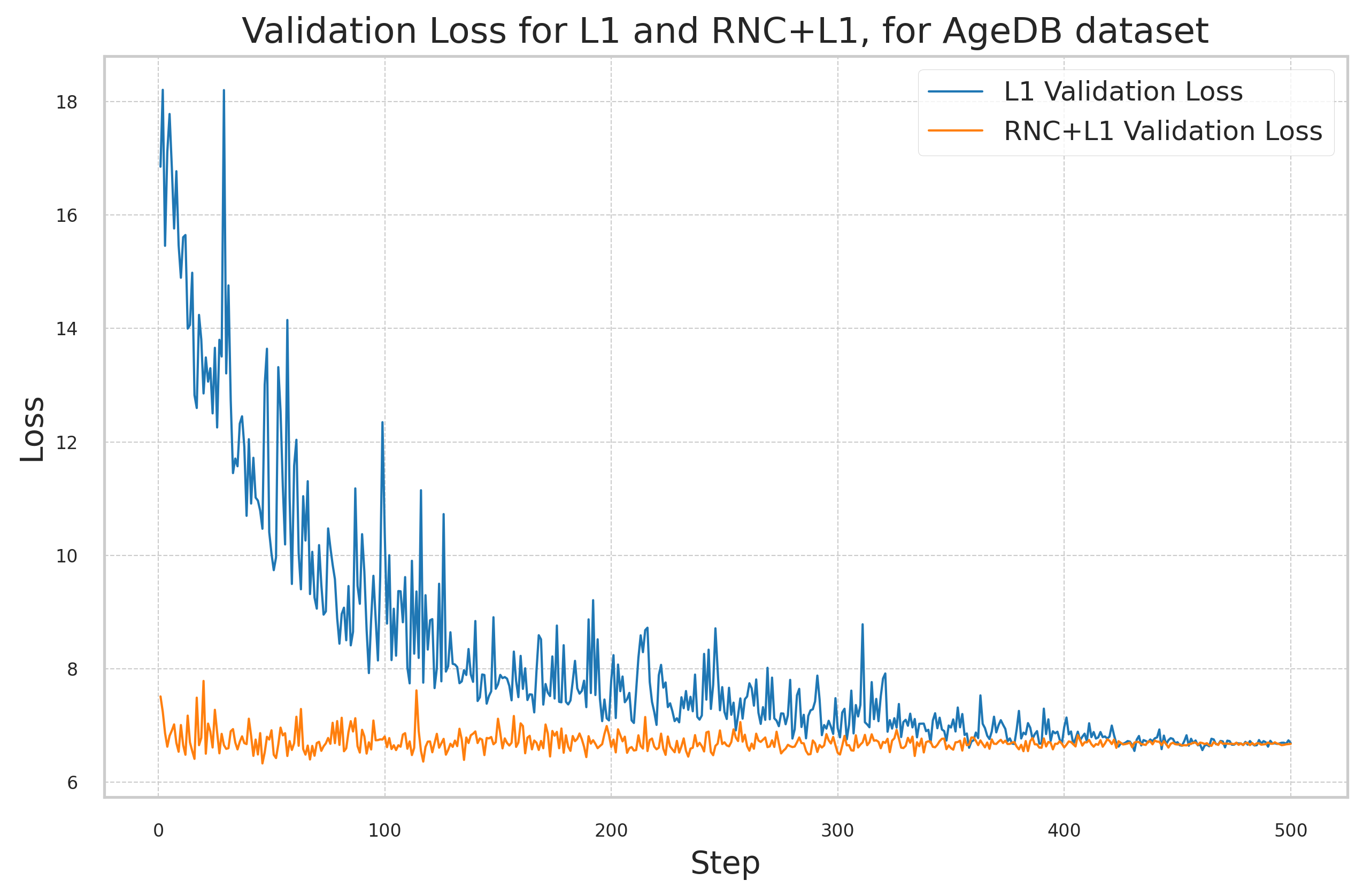}
\caption{Validation Loss for $L_1$ and $L_{RNC}+L_1$ on AgeDB dataset}
\label{fig2}
\end{figure}

On the validation data (Figure \ref{fig2}), $L_1$ shows a noisier but decreasing loss, which converges after 400–500 epochs. For the $L_{RNC}+L_1$ model, the validation loss starts at a very low value, even after only a few epochs: this observation is a sign that our model was able to learn the continuous representation in the embedding space, allowing it to make better predictions rapidly on unknown data. For the $L_1$ model, dealing with new data is more complex as the representation learned by the model is not necessarily continuous, making it harder to predict on new samples that do not resemble the previous ones.\\

We also conducted another experiment to test the robustness of our model: we created a different split of the data, still using \cite{moschoglou2017agedb}, by removing all images in the training set that correspond to an age ranging from 30 to 40. The validation set contains images that correspond to the complete range of ages, which should be a challenge for the standard $L_1$ model. By conducting this experiment, we wish to see if the continuity in the embedding space allows the model to perform better on such regression tasks, where the continuity within the images plays a key role.

\begin{figure}[htbp]
\centering
\includegraphics[width=3.5in]{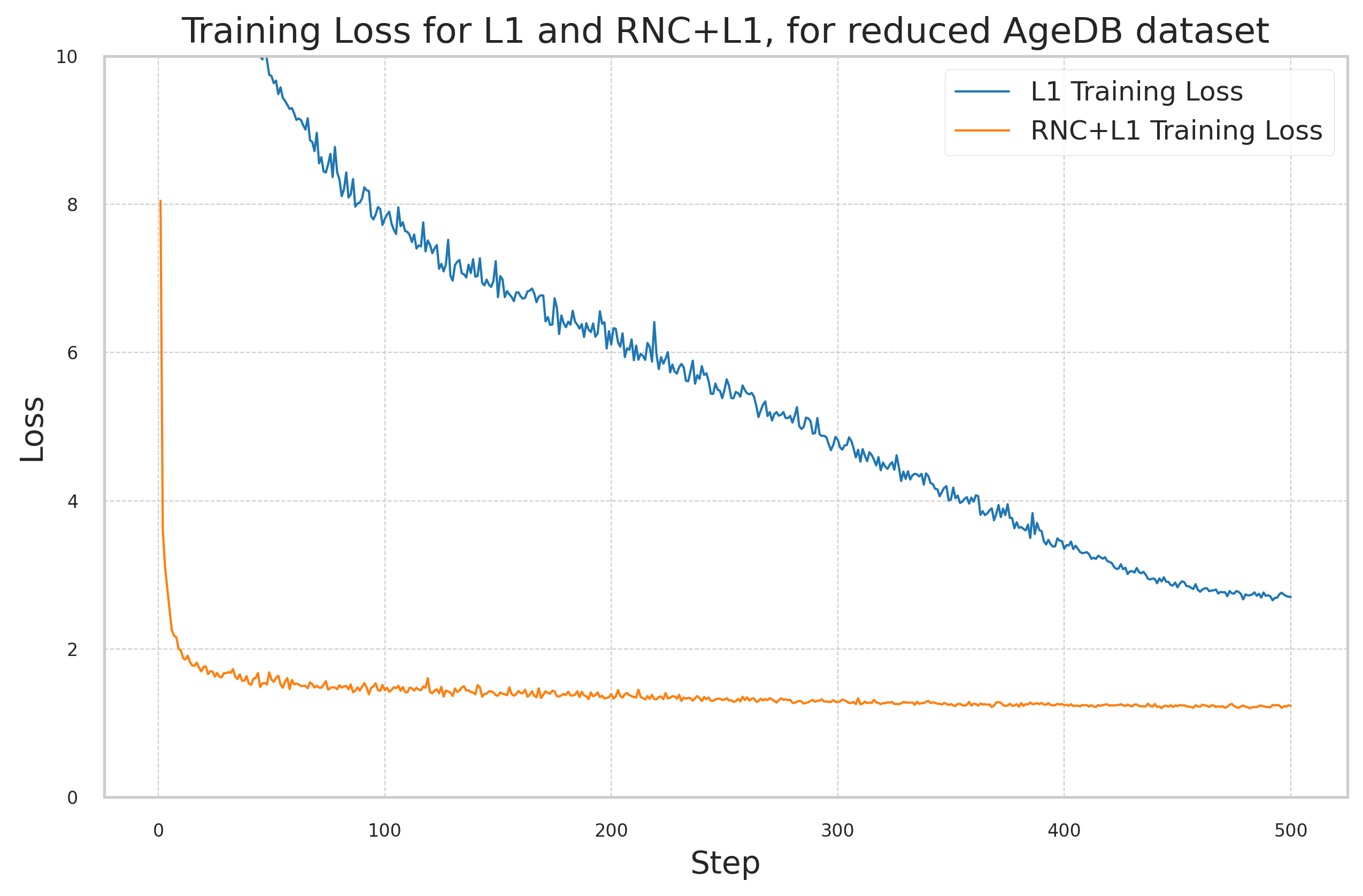}
\caption{Training Loss for $L_1$ and $L_{RNC}+L_1$ on modified AgeDB dataset}
\label{fig3}
\end{figure}

Figure \ref{fig3} shows the evolution of the training loss for both the $L_1$ and the $L_{RNC}+L_1$ models. We can note the high similarity between the learning curves on Figures \ref{fig1} and \ref{fig3}, which shows that the learning process is quite similar. We also see that the loss converges in both cases to a similar value. This observation serves as a good sanity check for our models: removing a certain range of ages from the training should be a challenge for the validation set, but should change the training process much. The interesting insights will appear when we consider the results for validation. \\

\newpage

\begin{figure}[htbp]
\centering
\includegraphics[width=3.5in]{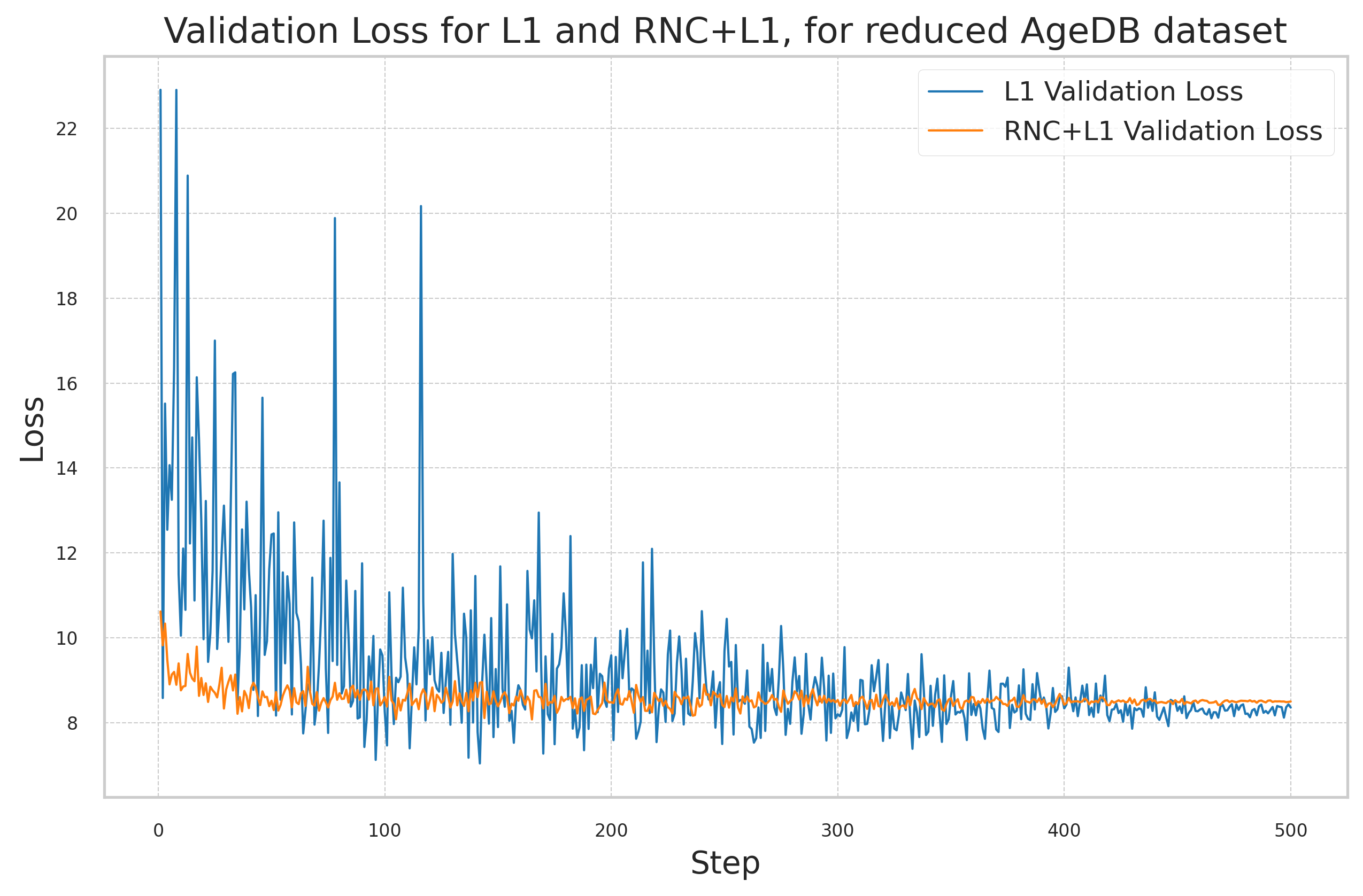}
\caption{Validation Loss for $L_1$ and $L_{RNC}+L_1$ on modified AgeDB dataset}
\label{fig4}
\end{figure}

Looking at Figure \ref{fig4}, we can observe important differences if we compare it with Figure \ref{fig2}. For the $L_1$ loss, the convergence is now much noisier and takes many more epochs. The final loss value to which it converges is also higher (from 6.75 to 8.5). The missing range of age in the training set makes it more difficult for the standard model to generalize to unseen data, which means making predictions in the missing range of data. For the $L_{RNC}+L_1$ model, Figure \ref{fig4} shows that we have a similar convergence process, although the final loss value is higher than with the regular split (from 6.75 to 8.5).\\

Overall, the new split affected the $L_1$ model much more than the $L_{RNC}+L_1$ for the speed of convergence ; but both models achieved a higher loss when dealing with a missing range in the training set.\\

The last experiment we conducted consisted in assessing the impact of training the encoder using $L_{RNC}$ on a new, smaller dataset. We used the dataset proposed in \cite{ismail2022estimating}, where the regression task is about estimating the weight of a mango from a single image of the fruit. The dataset contains 552 images, which significantly less than the previous dataset ; using less data for training will give us a hint to whether the $L_{RNC}+L_1$ model still performs better on smaller datasets. We used the same training process than for AgeDB (\cite{moschoglou2017agedb}), training both $L_1$ and $L_{RNC}+L_1$ models to compare performances.

\begin{figure}[htbp]
\centering
\includegraphics[width=3.5in]{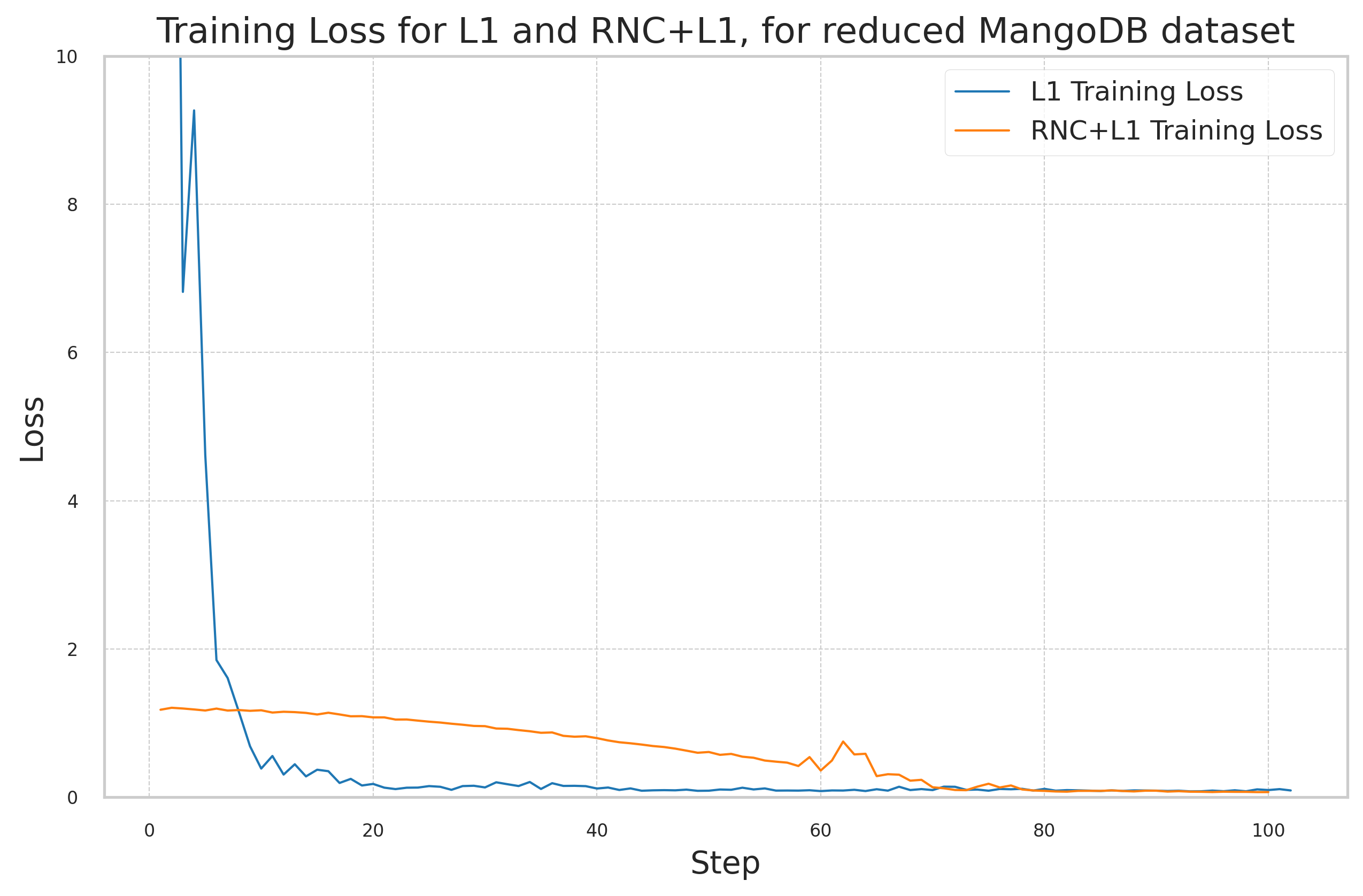}
\caption{Training Loss for $L_1$ and $L_{RNC}+L_1$ on MangoMassNet-552 dataset}
\label{fig5}
\end{figure}

\begin{figure}[htbp]
\centering
\includegraphics[width=3.5in]{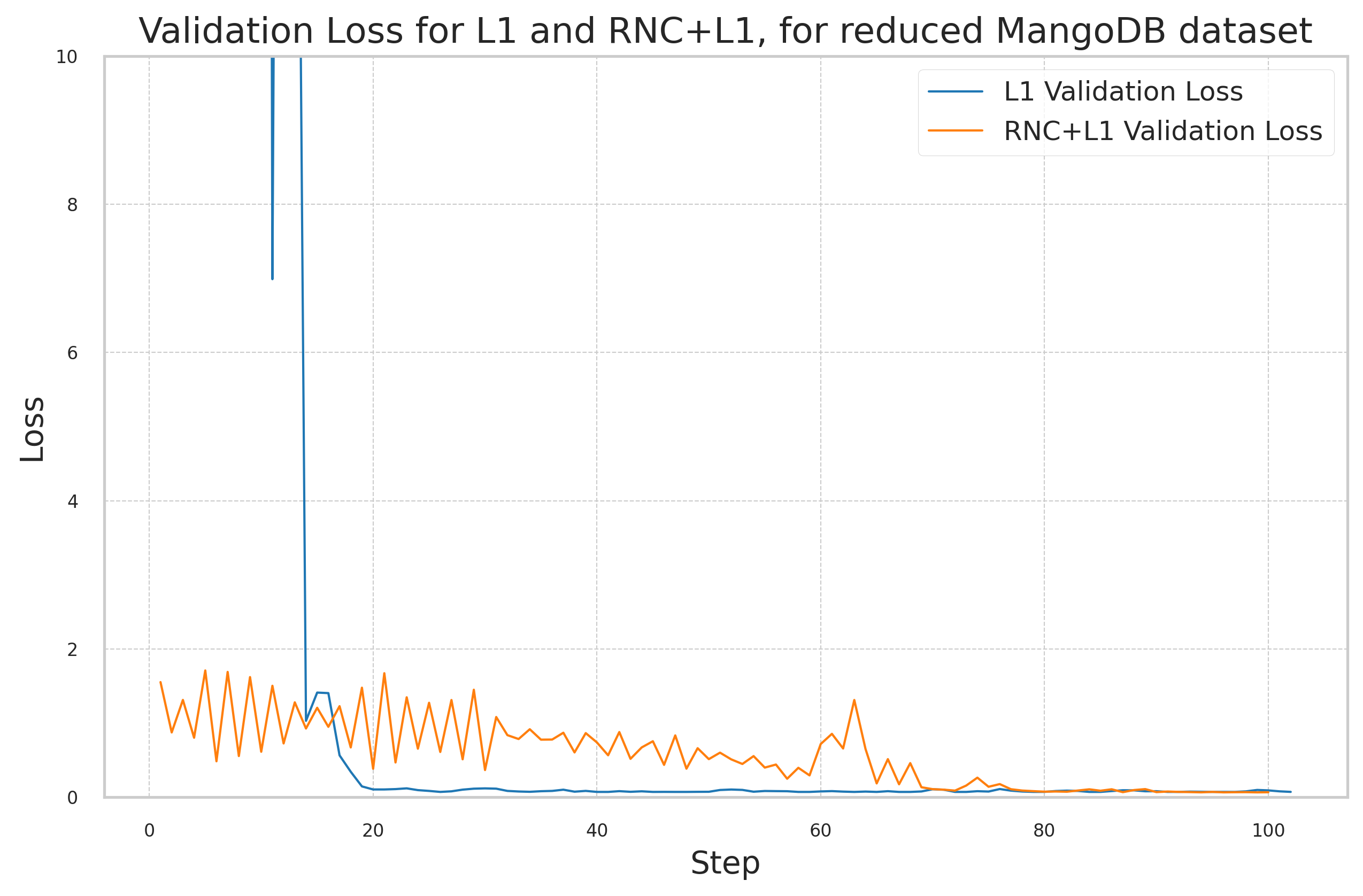}
\caption{Validation Loss for $L_1$ and $L_{RNC}+L_1$ on MangoMassNet-552 dataset}
\label{fig6}
\end{figure}

\newpage

Figures \ref{fig5} and \ref{fig6} clearly show us that both models converge faster, but the main difference with using a smaller dataset is that the $L_1$ model converges faster than $L_{RNC}+L_1$, for both training and validation. The surprising result could possibly come from the fact that the learner at the encoder is not learning as much as previously. Such observations are summarized in Table \ref{table1}, where we see that the final $L_{RNC}$ is higher when we reduce the size of the dataset.

\begin{table}[htbp]
\renewcommand{\arraystretch}{1.3}
\caption{Final RNC Loss Value at Encoder}
\label{table1}
\centering
\begin{tabular}{c|c|c}
\hline
\bfseries Experiment & \bfseries Final RNC Loss & \bfseries Dataset Size\\
\hline
AgeDB Normal Split & 3.92 & 16,488 \\
\hline
AgeDB Modified Split & 3.98 & 16,488\\
\hline
MangoMassNet & 5.37 & 552\\
\hline
\end{tabular}
\end{table}

\newpage

\section{Conclusion}

In this paper, we replicated the findings that were obtained in \cite{zha2024rank}, and we experimented deeper by testing the robustness of the model and using a different and smaller dataset. We demonstrated that, with enough data, using $L_{RNC}$ at the encoder can help to achieve better performance for regression tasks and make the model more robust to missing data in the training set.

\end{document}